# Learning the grammar of drug prescription: recurrent neural network grammars for medication information extraction in clinical texts

Ivan Lerner[1, 2, 3, 4], Jordan Jouffroy[1, 2, 3, 4], Anita Burgun[1, 2, 3, 4], and Antoine Neuraz[1, 2, 3, 4]

[1]Informatique biomédicale, CHU Necker, AP-HP, Paris, France

[2]INSERM UMR-S 872 Team 22, Information Sciences to support Personalized Medicine

[3]Sorbonne Paris Cité, Faculté de Médecine, Paris, France

[4]Université de Paris

Corresponding author: antoine.neuraz 'at' aphp.fr

## ABSTRACT

### Background and objective

In this study, we evaluated the RNNG, a neural top-down transition based parser, for medication information extraction in clinical texts.

### Methods

We evaluated this model on a French clinical corpus. The task was to extract the name of a drug (or a drug class), as well as attributes informing its administration: frequency, dosage, duration, condition and route of administration. We compared the RNNG model that jointly identifies entities, events and their relations with separate BiLSTMs models for entities, events and relations as baselines. We call seq-BiLSTMs the baseline models for relations extraction that takes as extra-input the output of the BiLSTMs for entities and events. Similarly, we evaluated seq-RNNG, a hybrid RNNG model that takes as extra-input the output of the BiLSTMs for entities and events.

**Results**

RNNG outperforms seq-BiLSTM for identifying complex relations, with on average 88.1 [84.4-91.6] % versus 69.9 [64.0-75.4] F-measure. However, RNNG tends to be weaker than the baseline BiLSTM on detecting entities, with on average 82.4 [80.8-83.8] versus 84.1 [82.7-85.6] % F- measure. RNNG trained only for detecting relations tends to be weaker than RNNG with the joint modelling objective, 87.4% [85.8-88.8] versus 88.5% [87.2-89.8]. Seq-RNNG is on par with BiLSTM for entities (84.0 [82.6-85.4] % F-measure) and with RNNG for relations (88.7 [87.4-90.0] % F-measure).

**Conclusion**

The performance of RNNG on relations can be explained both by the model architecture, which provides inductive bias to capture the hierarchy in the targets, and the joint modeling objective which allows the RNNG to learn richer representations. RNNG is efficient for modeling relations between entities or/and events in medical texts and its performances are close to those of a BiLSTM for entity and event detection.

## Introduction

Extracting relevant information from clinical text (e.g., drugs, symptoms, diagnosis ) is of the utmost importance for many applications in healthcare. This includes providing data for epidemiology, recruiting patients for clinical trials or any other tool to make clinical decisions more accurate and safe. If such information is often present in structured format, it may be incomplete (e.g., drug prescriptions omitting drugs prescribed outside the hospital) or not precise enough (e.g., diagnosis codes used for billing). For example, when the hospitalization of a patient is due to an adverse drug reaction, this information might be available only in free text in the medical record.

The era of neural models for named entity recognition (NER) and relation extraction has brought many improvements [1–4] in the last few years. These improvements were also driven by the use of rich contextual representation as inputs for supervised models, such as word embeddings [1,5] and more recently representations based on large pre-trained language models [6–8]. It has considerably reduced the need to develop rigid handcrafted rules for feature extraction. However, the cost of annotation for such systems, particularly when it comes to a high level of expertise, is still an important limitation [9,10].

Specifically, annotating relations is a difficult task. One way to lighten the burden of annotating relations is to use the dependencies between a given entity (e.g., "Aspirin") and other terms (e.g., "daily") that are dependent from the first entity [11]. The resulting representation is a tree, where "Aspirin" is called the parent and "daily" the child. More precisely, the sentence "(...) was given Aspirin 100 mg daily" is represented by one drug&attributes entity encapsulating the relations between 'Aspirin' and the dosage '100mg' and 'Aspirin' and its frequency 'daily' (in IOB: "O O B-D&A I-D&A I-D&A I-D&A"). Although this is not, strictly speaking, a *hierarchy,* we use the term "hierarchy" in the rest of the article since it denotes the graph of dependencies and the propagation of relations based on the links among entities.

This representation has shown to allow for a good coverage of all relations [12].

The most effective strategy to reduce the need for annotations or increase generalization performance in natural language processing (NLP) tasks is undoubtedly to use transfer learning from large pre-trained language models [6–8]. In addition, there is evidence that joint modeling or multi-task learning increases generalizability of systems trained on small datasets [13]. Indeed, features extracted from one task can benefit the other task. Such experiments have already proven successful in joint modeling of entities and relations [14]. Additionally, training one model for multiple tasks virtually diminishes the number of parameters for each task and the computational cost of the models. Another well investigated direction is to introduce inductive bias into the model, for instance by developing models tailored to exploit "natural" invariances in the data. In linguistics, according to the principle of compositionality, the interpretation of a complex expression is a function of the interpretation of its parts and the way they are assembled. This principle can be used as a strong inductive bias for developing NLP models [15]. Last but not least, extracting features from gazetteers or handcrafted rules remains effective in a low regime of annotated examples [10].

In this work, we developed a model for drug information extraction in French clinical texts. More precisely, the task consisted in extracting all information corresponding to two types of elements: the drug (or the drug class, e.g., anticoagulant) which is an entity and the administration or the prescription properties (referred to as attributes in the rest of the article), i.e., frequency, dosage, duration, condition and route of administration. The task also includes extracting other events related to a drug prescription such as *drug stopped* or *switch of a drug*. Of note, an attribute can be related to more than one drug, for instance the frequency (i.e.,"daily") in "Aspirin and Plavix, daily". This task is well suited to a hierarchical representation scheme, with a first layer of entities (including drug mentions, medication attributes and events), to which is added a second layer describing the relations between drug entities and their attributes, and finally a third layer to model the relations between a group of drug entities and their common attributes (see Figure 1).

The objective of this work was to develop a single model that would combine all the strategies mentioned above. We based our model on the recurrent neural network grammar (RNNG) [15]. Rationale for choosing RNNG was threefold: 1) the compatibility of RNNG with a hierarchical representation of labels, which comes with 2) the joint modeling of entities, events and relations and 3) an inductive bias for text data. Indeed, RNNG is a top-down transition based parser designed to model a sentence as a constituency tree, under the assumption that language can be modeled through the composition of nested constituents.

The transition from the constituency tree to the entities, events and relations modelisation objective was straightforward. Our main contributions are: 1) leverage transfer learning from large pre-trained language models, 2) articulation with external knowledge based features and task-specific transition rules and 3) custom modifications of the model detailed in the method's section "Modifications to the RNNG". We evaluated our model on a French dataset APmed [11], and compared it with a BiLSTM-CRF architecture. Finally, we report experiments shedding light on the impact of joint modeling of entities, events and relations for the RNNG, and on the complementarity of the two models, by investigating the performance of a sequential hybrid model seq-RNNG.

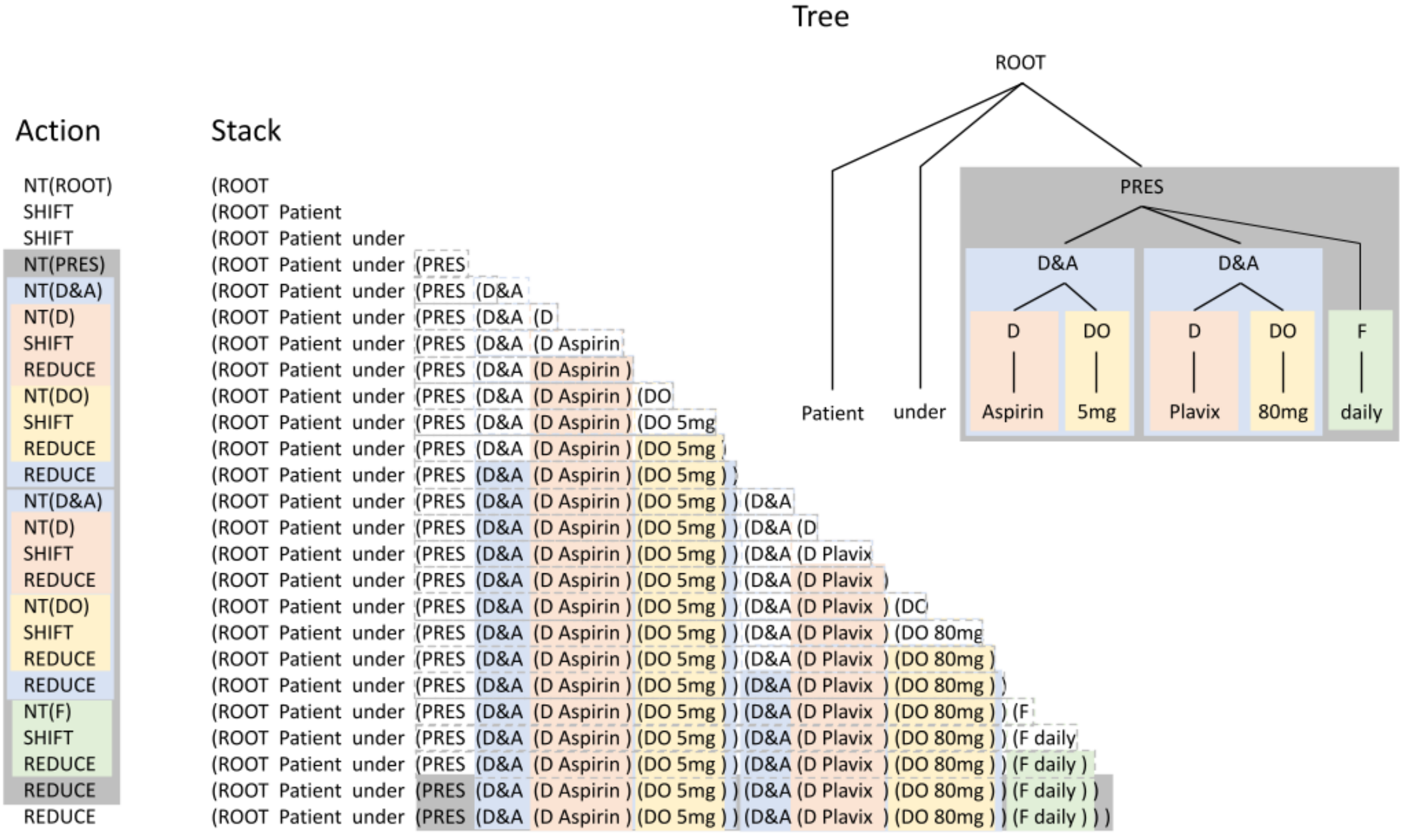

**Figure 1. RNNG allows representing entities and their relations in a hierarchical tree.**

D&A : *drug&attributes* relations; PRES : *prescription* relations. D : drug; DO : dose; F : frequency; NT : non terminal node. NT opens a new branch of the tree; SHIFT consumes tokens from the sequence and REDUCE closes the branch.

## Methods

### Dataset

| | Train | Dev | Test | Total | Median length [90%CI] |
|---|---|---|---|---|---|
| Documents | 203 | 37 | 80 | 320 | - |
| Tokens | 97529 | 16588 | 40058 | 154175 | - |
| **Relations** | | | | | |
| Drug&attributes | 814 | 128 | 290 | 1232 | 7 [3-18] |
| Prescription | 124 | 17 | 38 | 179 | 9 [3-26] |
| **Entities and attributes** | | | | | |
| Drug name | 1206 | 154 | 396 | 1756 | 1 [1-2] |
| Drug class | 226 | 33 | 76 | 335 | 1 [1-2] |
| Dose | 732 | 97 | 310 | 1139 | 2 [1-3] |
| Frequency | 523 | 61 | 176 | 760 | 2 [1-4] |
| Route | 76 | 17 | 54 | 147 | 1 [1-2] |
| Duration | 72 | 9 | 36 | 117 | 2 [1-2] |
| Condition | 61 | 4 | 28 | 93 | 3 [2-5] |
| **Events** | | | | | |
| Start | 136 | 20 | 50 | 206 | 1 [1-3] |
| Start & stop | 85 | 20 | 28 | 133 | 3 [1-5] |
| Stop | 72 | 18 | 29 | 119 | 3 [1-4] |
| Continue | 75 | 16 | 20 | 111 | 1 [1-2] |
| Switch | 22 | 5 | 12 | 39 | 1 [1-1] |
| Decrease | 18 | 2 | 3 | 23 | 1 [1-3] |
| Increase | 9 | 5 | 7 | 21 | 1 [1-1] |

**Table 1**: APmed corpus

We performed all the experiments on the APmed corpus [11], a corpus of clinical text reports in French, extracted from the AP-HP data warehouse [16]. APmed is a sample of 320 free text reports, including medical prescriptions, discharge reports, examinations, observation reports and emergency visits, randomly selected from the AP-HP clinical data-warehouse. APmed is built for drug information extraction, and includes drug name recognition, and their relation with the following attributes: frequency, dosage, duration, condition, route and events (drug stopped, switch of drug, etc.). This corpus also contains additional annotations such as temporality and certainty not included in this study. The corpus is divided into train, dev and test sets, summary statistics are reported in Table 1. Test set was selected according to the partition in [11], train and dev were splitted using stratified sampling based on the number of annotations by documents. We used a simple pre-processing of the text including lowercasing, sentence segmentation and word tokenization using custom regular expressions.

**Baseline model: (seq)-BiLSTM-CRF**

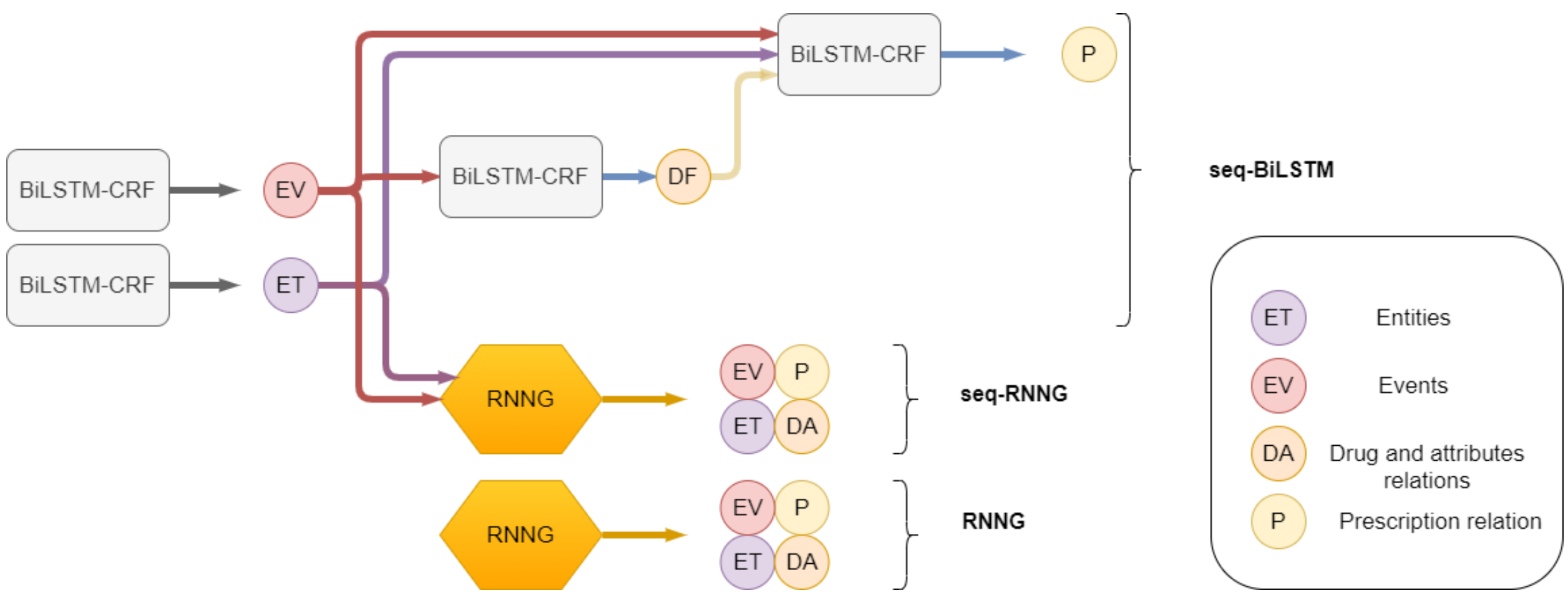

**Figure 2. Outputs of predicted entities as inputs for predicting relations**

Our baseline model is a bidirectional LSTM-CRF (BiLSTM-CRF) [1] architecture on top of contextual embeddings: ELMo embeddings [6] trained on a set of 100k clinical notes. The ELMo embeddings were constructed by concatenating the hidden representation of the convolutional character embedding layer and the two BiLSTM layers of the ELMo model. We used the flair implementation [17] for the LSTM-CRF model. We used word dropout which randomly drops an entire word representation, and locked dropout which randomly drops neurons with the same mask within a mini-batch, before and after the LSTM pass. Four different models were trained to extract (see Figure 2): 1) drug names and drug attributes, 2) drug events, 3) drug & attributes relations, and 4) prescription relations. As illustrated in Figure 2, the BIO labels of the models 1) and 2) were feeded as extra-features to the model 3), and the BIO labels of the models 1), 2) and 3) were feeded as extra-features to the model 4). These features were encoded by a categorical feature embedding of dimension 10, initialized randomly and learned during the task. These feature embeddings were then concatenated with the ELMo embeddings. We refer to the models 3) and 4) as seq-BiLSTM-CRF.

**The RNNG architecture**

The core architecture of our experiments is the RNNG in its discriminative configuration, which is a top-down transition based parser [15]. It was initially developed for language modeling and parsing. RNNG takes as input a sequence of tokens, and outputs a sequence of actions to build a tree. The tree structure is a graph composed of terminal (tokens) and non-terminal nodes (entities, events or relations). As illustrated in Figure 1, the RNNG can take three types of actions from the last open non-terminal node or root to form the tree: OPEN-NT appends a non-terminal node (*i.e.,* new branch of the tree), SHIFT appends a terminal node (*i.e.,* a new leaf of the tree), and REDUCE closes the last open non-terminal node (*i.e.,* closes a branch).

For each parsing step, the RNNG computes a latent vectorized representation of the tree using three structures: the buffer, the stack and the history of actions (Figure 1). Tokens and actions are both represented by embeddings. The role of the buffer is to compute a contextualized representation of the tokens in the sentence, each time that the RNNG takes a SHIFT action, a token is consumed and added into the stack. The stack computes a representation of the tree structure, by modelling a sequence containing actions embeddings, tokens representation from the buffer, and subtree representations computed through a composition function. The composition function can take the form of a simple summation or a BiLSTM. The history of actions computes a representation of past predicted actions. The buffer and history of actions are both LSTMs, and the stack architecture is a stack-LSTM [18]. A stack-LSTM is a LSTM whose access to the last hidden state is given by a pointing mechanism. The REDUCE action shifts the pointer of the stack-LSTM to the latent state before the opening of the last open-NT or ROOT. The latent representation of the tree is then the concatenation of the last hidden state of the three LSTMs structures (buffer, stack, history of actions).

Finally, the RNNG decodes this vectorized representation using a simple linear neural layer and a softmax. In the discriminative setting, the RNNG objective is to maximize the likelihood of the predicted tree, conditionally to the input sequence of tokens. Training is performed using oracles from the target tree. The predictions of the RNNG remains bounded by a set of transition rules. Transition rules can be defined *a priori* to constrain the list of actions available at each step depending on the current state of the tree. We based our developments on the open-source pyText implementation [19].

**Modifications to the RNNG**

Contextualized word embeddings

We tested the addition of a pre-trained ELMo model, to substitute contextualized embeddings to the classical word embedding representation. The ELMo model was pre-trained on a corpus of 100k clinical documents from a French hospital, with default parameters. The ELMo embeddings were constructed by concatenating the hidden representation of the convolutional character embedding layer and the two BiLSTM layers of the ELMo model, following the implementation from the library FLAIR [17] for comparison purposes.

Task-specific transition rules

We modified the constraints of the parser by introducing a set of task-specific transition rules, which combine flexibility in their definition and certification of a well-formed output tree. This set of rules can be generated automatically from the annotated train set and if necessary manually modified depending on the specific task. The possibility to automatically produce them from a small set of annotated examples could therefore be useful during an annotation process to help formalize the annotation guidelines. Indeed, the implementation of these rules allowed us to detect and correct a few annotation inconsistencies in the dataset. The rules were defined as follows: for each non-terminal, we constrain the set of children that can be opened, conditionally on its parent non-terminal, along with its maximum number. For instance, we can set that, the parent node being a drug & attributes relation, there is a limited set of possible non-terminals namely Drug, Frequency and Dosage,. We also set the maximum number of Drug nodes in a drug & attributes branch to 10, hence if the maximum is reached, the RNNG cannot open any new Drug non-terminal in the same branch of the tree. Similarly, a maximum number of SHIFT actions are assigned for each non-terminal node, thus controlling for the number of tokens for entities and relations.

Label smoothing

We modified the cross-entropy objective by introducing label smoothing, which soften the hard target of cross-entropy by introducing $\alpha$ such that for $K$ classes, $Y_k^{smoothed} = Y_k(1-\alpha) + \alpha/K$ [20]. The expected effect of label smoothing is to increase regularization and help calibrate the model.

Terminology-based features

We implemented a terminology-based feature (TBF) extractor, which matches tokens in a sentence to a set of terminologies. The terminology comes from a previous study on French clinical NER [10], and mostly includes UMLS® [21] and SNOMED 3.5 CT® terms [22], for Drug names, Sign and symptoms, Therapeutic procedures, Diagnostic procedures and Diseases. The best terminologies were selected during the optimization procedure as hyperparameters. The TBF outputs a sequence of BIO labels indicating the match with a given terminology. These categorical features are encoded in a 10 dimensional embedding, initialized randomly and learned during the NER task. This terminology embedding is then concatenated to the ELMo embedding. We used the open-source *flashtext* implementation - based on Trie dictionary data structure - for an efficient mapping from the terminologies [23].

Pre-labellisation of entities

To improve the performances of RNNG on the detection of entities, we added a BiLSTM-CRF layer before RNNG which predicted a BIO annotation for the entities. The BIO annotation, along with ELMo embeddings was then fed into the RNNG. We call this model seq-RNNG (Figure 3). In this setting, the RNNG could in theory leverage the predictions from the BiLSTM by learning predictive patterns of the BiLSTMs errors for detecting entities.

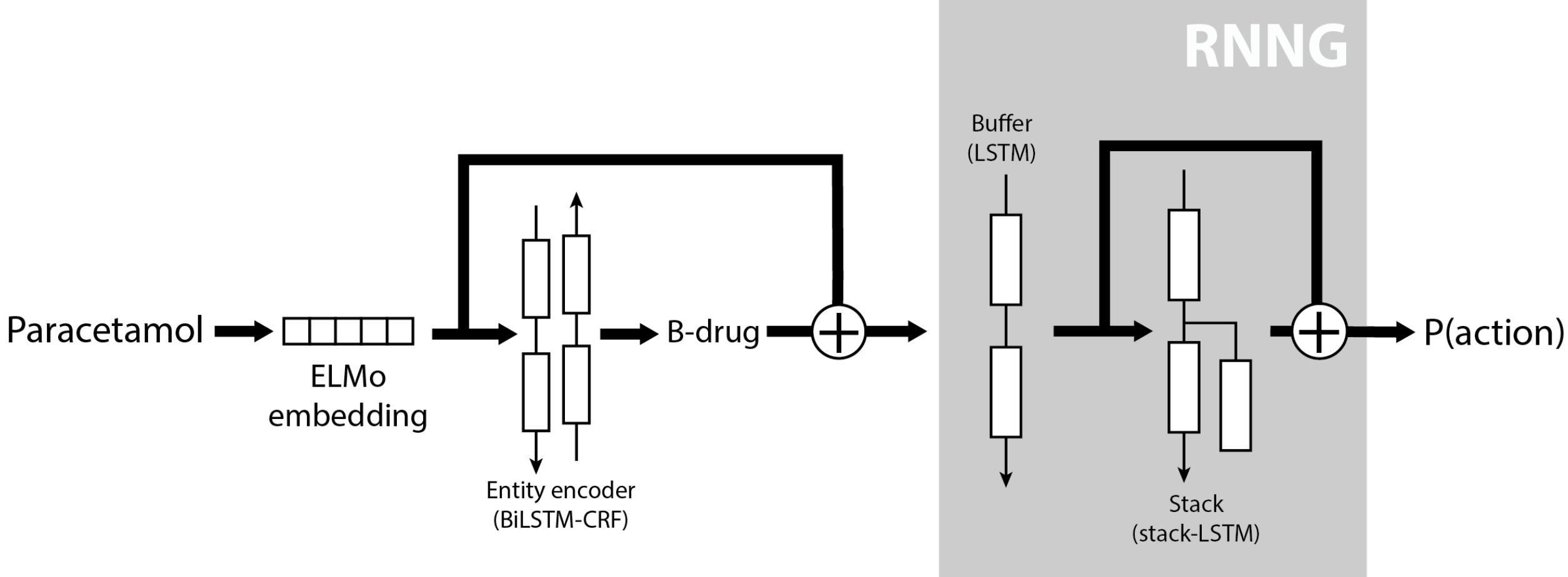

**Figure 3. Details of the seq-RNNG**

**Experiment on joint modeling of entities and relations**

We made experiments to explore the impact of joint modeling of entities and relations in the RNNG (Figure 2). We studied the performance of RNNG and seq-RNNG trained for relations extraction only. In this setting (see Table 4, ‘No joint model’), the RNNG does not compute composed representations of entities and events.

**Training and model selection**

We performed the hyperparameter tuning by exploring subspaces of the hyperparameter space using Bayesian optimization [24]. The initial experiments with the full training set were not satisfactory. Therefore,we chose to train the RNNG models on an undersampled version of the dataset with a 1:1 ratio between labeled and unlabeled sentences, which allowed us to retain only 10% of the unlabeled sentences. The parameter space explored included in addition to the modifications mentioned above: the number of layers and neurons for the BiLSTMs and RNNG, fastText [25] versus ELMo embeddings, dropout probability, L2-norm weight penalization, learning rate, optimizer type and use of a reduce on plateau learning rate scheduler (Table S2). All results of RNNG are presented with beam-search at inference (beam size = 3).

**Metrics**

We compared the models using precision (P), Recall (R) and F-measure (F1).

We used an exact-match NER definition for entities and events, where an instance is defined as the group of tokens that make up an entity. An instance is counted as true positive (TP) if all the predicted tokens match the ground truth annotation, else as false positive (FP) or false negative (FN).

For relations, we define an instance as a pair of a drug (or class) name and an attribute or event. A relation instance is counted as TP if correctly labeled by a model, else FP or FN. We can compute the relations metrics based on the predicted entities or events of the model. This will reflect the actual performance of the model. Another way is to base the metrics on the ground truth entities or events (oracles). This allows us to assess the quality of the relations labels independently of the performance of the models on entities or events. For a set of entities, events or relation instances, P, R and F1 are defined as

$$R = \frac{TP}{TP + FN},\ P = \frac{TP}{TP + FP}\ and\ F1 = 2.\frac{P.R}{P + R}$$

We report the mean and 90% confidence interval (5 and 95 percentiles) by bootstrapping the instances of the test set with 1,000 iterations. Note that this only accounts for uncertainty of the model performance on this dataset, and not uncertainty of the training procedure. In tables, results are reported for relations, entities and events, and for each label from most common to least common in the dataset. Finally, we performed an error analysis for prescription relations, qualitatively by analyzing examples where seq-BiLSTM and RNNG predicted different relations, and quantitatively by comparing performance for long and short *prescription* relations (length of the meta-entity prescription superior or equal to the median number of tokens).

## Results

### RNNG compared to (seq)-BiLSTM-CRF

| | **(seq)-BiLSTM** | **RNNG** | **seq-RNNG** |
|---|---|---|---|
| **Relations** | 84.6 [83.1-86.1] | 88.5 [87.2-89.8] | 88.7 [87.4-90.0] |
| **Entities** | 84.1 [82.7-85.6] | 82.4 [80.8-83.8] | 84.0 [82.6-85.4] |
| **Events** | 53.0 [46.3-59.3] | 49.7 [43.4-55.9] | 52.5 [46.5-59.0] |

**Table 2**: RNNG outperforms seq-BiLSTM-CRF for identifying relations.

Micro-average F-measure and its boostrapped 5 and 95 percentiles as computed on the test set of APmed.

We present in Table 2 the aggregated results of RNNG performance, in comparison with the (seq)-BiLSTM-CRF (see Table 3 for detailed results). The RNNG tends to outperform the seq-BiLSTM-CRF on detecting relations, with F measures 88.5 [87.2-89.8] % versus 84.6 [83.1-86.1] %, respectively. This difference was significant in the deeper hierarchy of prescription, 88.1 [84.4-91.6] % versus 69.9 [64.0-75.4] of F-measure, respectively. However, the RNNG achieved lower performances than the BiLSTM-CRF on the extraction of entities, with F-measures of 82.4 [80.8-83.8] % versus 84.1 [82.7-85.6] %, respectively. It also shows lower performances on the extractions of events. The seq-RNNG performs similarly on entities and events than the BiLSTM-CRF alone, and similarly on relations than RNNG, providing the overall best results.

| | (seq)-BiLSTM | RNNG | seq-RNNG | Support Train | Support test |
|---|---|---|---|---|---|
| **Relations** | | | | | |
| drug&attributes | 87.0 [85.5-88.4] | 88.6 [87.2-90.0] | 88.6 [87.1-89.8] | 814 | 290 |
| prescription | 69.9 [64.0-75.4] | 88.1 [84.4-91.6] | 89.2 [85.9-92.3] | 124 | 38 |
| **Entities and attributes** | | | | | |
| Drug name | 91.5 [89.8-93.2] | 87.4 [85.3-89.4] | 91.3 [89.6-92.9] | 1206 | 396 |
| Dose | 88.7 [86.3-90.9] | 85.4 [83.0-87.7] | 86.7 [84.3-89.0] | 732 | 310 |
| Frequency | 80.1 [76.1-83.8] | 85.2 [81.9-88.5] | 81.0 [77.3-84.7] | 523 | 176 |
| Drug class | 57.2 [49.1-64.3] | 63.1 [55.8-70.1] | 61.2 [53.7-68.6] | 226 | 76 |
| Route | 66.4 [58.1-74.2] | 76.3 [68.9-83.6] | 67.4 [58.7-75.4] | 76 | 54 |
| Duration | 85.0 [76.7-91.8] | 61.9 [49.2-74.0] | 83.3 [74.6-90.4] | 72 | 36 |
| Condition | 62.6 [50.0-75.0] | 52.3 [36.3-66.7] | 63.9 [50.9-75.0] | 61 | 28 |
| **Events** | | | | | |
| Start | 58.5 [47.4-68.2] | 49.4 [38.5-59.5] | 56.6 [46.2-66.7] | 136 | 50 |
| Start & stop | 42.9 [27.3-57.6] | 26.2 [13.0-40.0] | 37.0 [22.2-52.4] | 85 | 28 |
| Stop | 63.7 [51.2-75.5] | 66.4 [54.5-77.1] | 61.9 [48.9-73.5] | 72 | 29 |
| Continue | 50.9 [35.3-66.7] | 67.1 [51.2-80.0] | 53.8 [37.8-68.4] | 75 | 20 |
| Switch | 36.9 [11.8-62.5] | 13.6 [0.0-34.8] | 36.0 [0.0-62.5] | 22 | 12 |
| Decrease | 0.0 [0.0-0.0] | 0.0 [0.0-0.0] | 34.9 [0.0-80.0] | 18 | 3 |
| Increase | 36.8 [0.0-71.4] | 52.1 [16.7-82.4] | 41.3 [0.0-75.0] | 9 | 7 |

**Table 3**: Details of RNNG and (seq)-BiLSTM performances.

Micro-average F-measure and its boostrapped 5 and 95 percentiles as computed on the test set of APmed.

**The impact of jointly modeling entities and relations**

| | **Joint model** | **Relations** | **Entities** | **Events** |
|---|---|---|---|---|
| **(seq)-BiLSTM-CRF** | No | 84.6 [83.1-86.1] | 84.1 [82.7-85.6] | 53.0 [46.3-59.3] |
| **RNNG** | Yes | 88.5 [87.2-89.8] | 82.4 [80.8-83.8] | 49.7 [43.4-55.9] |
| **RNNG** | No | 87.4 [85.8-88.8] | - | - |
| **seq-RNNG** | No | 87.3 [85.8-88.8] | - | - |
| **seq-RNNG** | Yes | 88.7 [87.4-90.0] | 84.0 [82.6-85.4] | 52.5 [46.5-59.0] |

**Table 4**: Joint modelling helps RNNG with relations extractions.

Micro-average F-measure and its boostrapped 5 and 95 percentiles as computed on the test set of APmed. seq-RNNG uses the prediction of the BiLSTM models for entities and events as features. RNNG (or seq-RNNG) is trained for extracting relations only ('No joint model'), or jointly relations, entities and events ('Yes joint model'). In the 'No joint model' setting, RNNG (or seq-RNNG) does not compute composed representations of entities and events.

The results of the joint modeling experiments are presented in Table 4. The RNNG trained for jointly modeling relations, entities, and events outperforms the RNNG trained only for relations, 88.5 [87.2-89.8] % versus 87.4 [85.8-88.8] % F-measure for detecting relations, respectively. This difference remains even when providing RNNG with information about the entities from the BiLSTM-CRF (seq-RNNG), with 88.5 [87.2-89.8] % versus 87.3 [85.8-88.8] %, respectively.

**Error analysis**

We provide the reader with three interesting examples where the RNNG and seq-BiLSTM perform differently. First, in "Probabilistic antibiotic therapy by TAZOCILLINE and VANCOMYCINE J0 on 30 08 2012" (in french: "Antibiotherapie probabiliste par TAZOCILLINE et VANCOMYCINE J0 le 30 08 2012"), the RNNG gets the relation between both antibiotherapy and their date of initiation, whereas the seq-BILSTM does not. Second, in "Bi anti-aggregation by Kardegic + Plavix for 1 month, then treatment by Kardegic 160 mg alone" (in french: "Bi antiaggregation par Kardegic + Plavix pendant 1 mois, puis traitement par Kardegic 160 mg"), the RNNG gets the relation between the two stage of anti-aggregation therapy "Kardegic + Plavix" and "Kardegic", in relation by a switch event (word "then"), that the seq-BILSTM doesn't get. Third, in "prescription of amoxicillin, paracetamol codeine and paroex" (in french: "prescription d amoxicilline , paracetamol codeine et paroex"), RNNG fails to discover the relation between the start event ("prescription of") and the three drugs, whereas the seq-BILSTM does.

In addition, we found that RNNG outperforms seq-BiLSTM by a larger margin for long distance relations (72.5 [66.7-78.2] % versus 93.2 [90.1-96.0] % F-measure) than short distance relations (55.5 [38.9-70.6] % versus 61.3 [45.0-75.6] % F-measure).

**Discussion**

In this study, we evaluated models on the task of extracting medication information, which are drug names and attributes or events and how they are related. We show that RNNG, an LSTM based architecture developed to explicitly compose a hierarchical representation of a sentence, outperforms a simpler BiLSTM-CRF baseline to model complex relations. Furthermore, the sequential combination of BiLSTM-CRF and RNNG (seq-RNNG) achieved better overall performances than RNNG and BiLSTM-CRF alone.

**RNNG for relations and BiLSTM-CRF for entities**

RNNG outperforms seq-BiLSTM-CRF for detecting relations (see Table 3), especially the higher hierarchy of prescriptions. This performance is in part explained by the joint modeling objective of RNNG (see Table 4), which allows the RNNG to explicitly compose entities and events representations. Indeed, feeding naively information related to frontiers of entities and events as a feature in seq-RNNG, does not allow to match the performance of RNNG or seq-RNNG trained with the joint modeling objective. RNNG achieves lower performances than the BiLSTM-CRF on entities. The small effective size of events makes the comparison uncertain, but it seems that RNNG is also weaker on detecting events.

The underperformance is mitigated as the RNNG has still higher performances for certain classes (*e.g.,* Frequency, Drug class and Route). However, these discrepancies between entity or event classes does not seem associated with the size of the entities, nor the number of training examples.

One of the assumptions we made with respect to the seq-RNNG was that the RNNG part would be able to outperform the BiLSTM-CRF predictions on entities and events. Indeed, if these two models were to make different types of errors, the RNNG could then learn predictive patterns of the BiLSTM errors and reach better performance.

Another objective of seq-RNNG was to study the possibility of a three-headed model, with one common contextual embeddings, two BiLSTM-CRF heads for the entities and events, and one RNNG head for relations. Nevertheless, the RNNG without the joint modeling objective revealed lower performances, possibly because it does not compose the entities in a single representation. Therefore, a future direction to make such a model effective would be to integrate BiLSTM predictions into the RNNG architecture, for example by providing them as oracle actions to be taken instead of additional features.

Note that the results presented for relations are computed using oracle entities to enable comparisons across models. The actual performance for relations using the predicted entities of each models are much lower (see Table S1), with on average 65.4 [63.2-67.5] % and 64.5 [62.4-66.6] % F-measure for seq-BiLSTM-CRF and RNNG, respectively.

**Limitations**

First, the hierarchical representation of drug prescription is not suited for representing all types of relations. For instance, relations between entities across sentences, or overlapping relations between entities such as in "Aspirin and Plavix (80mg and 20 mg)". However, such settings are rare enough so that it did not occur in the annotated dataset.

Second, the RNNG model only accepts a mini-batch size of 1, which does not allow the full potential of parallelization of modern GPUs. In the same manner, the model used in this work is designed to compute representations of single sentences. Therefore, relations spreading over multiple sentences would not fit. To enable multi-sentence representations one would have to add a context element to the model such as initializing the hidden states of the LTSMs with the last hidden states of the previous sentence or concatenating previous and current hidden states.

Another limitation is that we did not account for the training procedure uncertainty, caused by the convergence of the models to different local minimas, by running several runs with the same parameters. We assumed that the variance arising from the sampling of documents, accounted for by the bootstrapping procedure when computing the metrics, would be much larger. This choice was also motivated by the cost of training models in terms of time and computation resources. Finally, we did not use Transformer architectures (e.g., BERT) for this work. Although it might lead to better performances, it was not feasible during the time of this study to train a BERT model for French clinical text. Therefore, we decided to use an ELMo model trained on French clinical text.

**Related work**

In FABLE [26], the authors introduced - for drug prescription extraction - the treatment of relations extraction as a sequence labeling problem, with relations as parent entities of children nested entities.

They modeled entities and relations sequentially, first by predicting entities, then by using the predicted entities and context token embeddings as inputs to model relations, alongside with a distance-to-entity feature and a word shape feature. Their best model using a semi-supervised approach reached 87.8% F-measure, close to the former work of Patrick et al. [27] which reached 88.16% F-measure.

Our work is closely inspired by Gupta et al. [12], who experimented with the same model in a slot filling task. They showed that RNNG outperformed other sequence to sequence architecture, based on LSTMs, CNNs or transformers. As the tree hierarchy for our task was limited to 3, we were able to compare with the seq-BiLSTM-CRF, which is more suited for a sequence labeling task than a seq-2-seq architecture. In addition, it revealed that the gain from using the RNNG model compared to a BiLSTM was proportional to the depth of the tree. In Wang et al. [2], the authors used a bottom-up transition-based parser, and found on the GENIA corpus [28] that their system outperforms other systems only for nested entities. On non-nested entities, their system is outperformed by Ju et al. [3], which for non-nested entities is equivalent to a BiLSTM-CRF. These results are consistent with ours where the bottom of the hierarchy, entities and events, did not benefit from the RNNG. They are also in line with the design of the RNNG architecture, within which the stack-LSTM allows to create shortcuts between representations of distant entities or events in the same sentence. Indeed, by learning to compose a single representation from multiple representations in a sublevel of the hierarchy (e.g., a single representation of a multi-tokens entity), RNNG is diminishing the sequence length at upper levels, hence decreasing the complexity of the function to learn. An interesting experimentation would include a layer of labels below entities and events in the hierarchy, such as POS tags. In addition, including the dependency structure when composing RNNG representations is likely to improve the results, as pioneering work on the extraction of prescription information has shown that a rule-based system exploiting the dependency structure achieves a high degree of accuracy on this task [29,30].

Another direction of research to model hierarchical outputs could leverage graph-based models [31–33]. In this area, deep biaffine attention [34], initially developed for dependency parsing, has revealed interesting results with about 1.5% gain compared to a linear transformation for modeling hierarchical structure of entities and relations [4].

In the medical domain, the combination of expert knowledge and distributed representation (embeddings) has proven efficient [35–37] to increase the performance of a BiLSTM-CRF [11], for instance reaching 94.24 [94.22-94.27] % F-measure on drug recognition.

## Conclusion

In this study, we experimented with RNNG, a neural top-down transition based parser, for the joint modeling of entities and relations of drug prescriptions in clinical texts. We showed evidence that the RNNG architecture and the joint modeling of entities and relations improved performance for relations detection.

## Declarations

### Authors' contributions

IL contributed to conceptualization; data curation; formal analysis; methodology; software;

writing - original draft. JJ contributed to conceptualization; data curation; writing - review & editing. AN contributed to conceptualization; data curation; formal analysis; software ; resources; supervision; validation; writing - original draft; writing - review & editing. AB contributed to project administration; resources; supervision; validation; writing - review & editing.

### Declarations of interests

The author declares no conflicting interests.

### Acknowledgments

We thank Sarah Feldman who took part in the annotation of the APmed dataset.

**Data and code**

The corpus can be made available on the condition that a research project is accepted by the scientific and ethics committee of the AP-HP health data warehouse (http://recherche.aphp.fr/eds/recherche/). Code for the modified RNNG model is available at https://gitlab.com/lerner.ivan/medrnng.

**Funding**

The authors received no specific funding for this work.

**Ethics**

This project was accepted by the scientific and ethics committee of the AP-HP health data warehouse as CSE-18-0025.

## Supplementary

**Table S1.** Performance of relations extractions including entities prediction errors.

| | **seq-BiLSTM** | **RNNG** | **seq-RNNG** |
|---|---|---|---|
| **drug&attributes** | 70.0 [67.7-72.3] | 70.1 [68.0-72.1] | 69.2 [66.9-71.4] |
| **prescription** | 29.2 [22.0-36.8] | 30.8 [24.5-37.1] | 40.6 [34.7-46.5] |
| **micro-average** | 65.4 [63.2-67.5] | 64.5 [62.4-66.6] | 64.5 [62.4-66.6] |

Micro-average F-measure and its boostrapped 5 and 95 percentiles as computed on the test set of APmed.

**Table S2.** Best hyperparameters for RNNG

| Group | Hyperparameter | Value |
|---|---|---|
| **Architecture** | | |
| | # hidden units per LSTM layer | 350 |
| | Number of layers | 2 |
| | Bidirectionnal LSTM | True |
| | Embedding type | Elmo |
| | Use task specific transition rule | True |
| | Composition function type | LSTM-based |
| | Use stack | True |
| | Use action | True |
| | Use buffer | True |
| **Regularization** | | |
| | LSTM Dropout | 0.46 |
| | Embedding dropout | 0.16 |
| | Label smoothing | 0.17 |
| **Training** | | |
| ADAM optimizer | learning rate | 0.02 |
| | L2 norm | 5.10^-6 |
| Reduce on plateau scheduler | min learning rate | 1.10^-4 |
| | factor | 0.1 |
| Early stopping | patience | 20 |
| Gradient clipping | | 1 |